\documentclass[conference]{IEEEtran}
\def\IEEEtitletopspace{18pt}
\IEEEoverridecommandlockouts

\usepackage[ruled,vlined]{algorithm2e}
\usepackage{blindtext}
\usepackage{graphicx}
\usepackage[utf8]{inputenc}
\usepackage{mathtools}
\usepackage{amsmath, amsthm, amsfonts, amssymb,color}
\usepackage[font=footnotesize,labelformat=simple]{subcaption}

\usepackage{tabu}

\usepackage{epsfig} 
\usepackage{psfrag}
\usepackage{mathrsfs}

\usepackage{multirow}
\usepackage{array,booktabs,arydshln,xcolor}
\usepackage{soul}
\usepackage{mathtools}
\usepackage{nccmath}
\usepackage{caption}
\usepackage{nccmath}
\usepackage{comment}
\usepackage{pdfpages}

\usepackage{float}
\usepackage{graphicx}
\usepackage{epstopdf}
\usepackage{color}
\usepackage{verbatim}
\usepackage{tikz,times}
\usepackage{xcolor}
\usepackage{lipsum}
\usepackage{makecell}
\usepackage{siunitx}
\usepackage{stfloats}
\usepackage[colorlinks=true,allcolors = cyan,citecolor=cyan]{hyperref}
\usepackage{tikz, cite}
\usepackage{dsfont}
\usepackage{orcidlink}
\usepackage{svg}

\newcommand\copyrighttext{%
  \footnotesize \textcopyright \the\year{} IEEE. Personal use of this material is permitted. Permission from IEEE must be obtained for all other uses, including reprinting/republishing this material for advertising or promotional purposes, collecting new collected works for resale or redistribution to servers or lists, or reuse of any copyrighted component of this work in other works.}

\newcommand\copyrightnotice{%
\begin{tikzpicture}[remember picture,overlay]
\node[anchor=south,yshift=10pt] at (current page.south) {\fbox{\parbox{\dimexpr0.85\textwidth-\fboxsep-\fboxrule\relax}{\copyrighttext}}};
\end{tikzpicture}%
}

\title{\LARGE \bf
Learning High-Level Decision Making with an Interaction-Aware Attention-Based Network in Autonomous Driving 
}

\author{Marcelo Contreras\,\orcidlink{0009-0005-4452-605X}$^{1}$, Willi Poh\,\orcidlink{0009-0007-8217-6438}$^{2}$, Christoph Stiller\,\orcidlink{0000-0003-4165-2075}$^{2}$, Ehsan Hashemi\,\orcidlink{0000-0002-6236-7516}$^{1}$, 
\thanks{$^{1}$ M. Contreras and E. Hashemi (Corresponding Author, {\tt\small ehashemi@ualberta.ca}) are with the NODE lab, University of Alberta, Edmonton, AB, T6G 1H9 Canada.}
\thanks{$^{2}$ W. Poh and C. Stiller are with the MRT Institute, Karlsruhe Institute of Technology, Karlsruhe, Germany.}
}

\begin{document}

\maketitle
\copyrightnotice
\thispagestyle{empty}
\pagestyle{empty}

\begin{abstract}
Reliable learning-based high-level decision making for lane changes and speed control in automated driving must accommodate dynamically sized inputs due to varying scene traffic flow. DeepSet and its variants represent the state of the art among shared-encoder approaches; however, they neglect explicit traffic interaction modeling, limiting performance in negotiation-intensive scenarios such as intersections. Attention-based methods capture interactions among static and dynamic agents, but incur quadratic memory and computational complexity and provide limited control over representation granularity. Inspired by Perceiver IO, an attention-based architecture, DecisionPerceiver, is proposed to project dynamic agent features into a fixed-size latent space, where feature granularity is regulated by the number of latent queries, improving scalability for larger networks. A finer discretization of the action set is further proposed to increase the performance gain due to interaction awareness. Extensive evaluations across three driving scenarios that require different levels of interaction awareness demonstrate consistent performance gains and generalization across various navigation objectives. In addition, the proposed architecture is assessed in scenarios with an increasing number of vehicles to demonstrate scalability. 
\end{abstract}
\section{Introduction} \label{sec:introduction}
Reinforcement learning (RL) has emerged as a powerful framework for autonomous driving policy learning \cite{rl1,rl2,rl3}, as it optimizes decision-making through closed-loop interaction with the environment. In contrast to imitation learning (IL), which is limited by the distribution and coverage of expert demonstrations \cite{lu2023imitationenoughrobustifyingimitation}, RL enables performance beyond demonstrated behaviors and improved generalization to novel scenarios. A key challenge in neural policy representations is sensitivity to input ordering, as the number and arrangement of surrounding agents vary dynamically with perception outputs. Prior works address this using convolutional neural networks (CNNs) with grid-based or bird’s-eye view (BEV) representations centered one the ego vehicle \cite{chen2019modelfreedeepreinforcementlearning}. While effective for spatial encoding, such inputs introduce redundancy and may obscure explicit kinematic states critical for decision making.

To address permutation sensitivity, the DeepSets architecture \cite{hugle_dynamic_2019} was introduced to process variable-size inputs using shared feature encoders combined with permutation-invariant aggregation operators, such as pooling or summation. This formulation has been extended for long-horizon reasoning \cite{kalweit_q-learning_2021}, surrogate-objective training \cite{kalweit_deep_2022}, explainable policies \cite{schier_explainable_2025}, and mitigation of multilayer perceptrons' spectral bias via Fourier feature mappings \cite{schier_deep_2023}.

\begin{figure}
    \centering
    \includegraphics[width=0.7\linewidth]{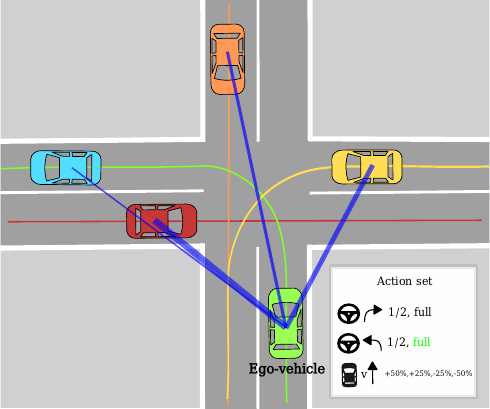}
    \caption{Traffic-participant interactions are modeled by projecting agent features into a latent space. Based on agents’ kinematic states and inferred interaction strength, the ego vehicle selects a full left-lane-change maneuver. Blue edges depict ego-to-agent interactions, with thickness proportional to intensity. The discrete action set uses finer lane-change and speed-adjustment discretization to improve trajectory tracking.}
    \label{fig:overview}
\end{figure}

DeepSets exhibits a key limitation: feature collapse after aggregating shared encoder outputs into a single vector, which restricts explicit modeling of inter-agent interactions essential for negotiation scenarios. To capture such interactions, graph neural networks (GNNs) \cite{huegle_dynamic_2020, peiss_graph-based_2023, schier_deep_2023} model vehicles as nodes and interaction strengths as learned edges, enabling structured relational reasoning. These approaches can also incorporate traffic topology and rules \cite{konstantinidis_modeling_2023}. However, GNNs rely on localized node updates and face optimization and batching challenges when scaling to large numbers of agents. Transformers and attention-based methods are widely adopted in natural language processing \cite{vaswani2023attentionneed} and computer vision \cite{trockman2022patchesneed} due to their ability to process long sequences and capture global dependencies. Several approaches have been explored in autonomous driving tasks \cite{leurent_social_2019,han_spformer_2024,qiu_human-like_2024}, including GNNs, where attention-based variants have in some cases outperformed their graph counterparts. However, attention mechanisms incur quadratic computational complexity with respect to input size, limiting real-time deployment.

An attention-based architecture, DecisionPerceiver, is therefore proposed to improve computational efficiency while preserving interaction modeling capacity. Inspired by Perceiver-IO \cite{jaegle_perceiver_2022}, the network projects variable-size inputs into a fixed-dimensional latent space, decoupling computational cost from input size while retaining interaction awareness among traffic agents (see Fig.~\ref{fig:overview}). The model is trained by RL to derive effective behavioral planning policies that reduce collision rates while maximizing speed and comfort. Furthermore, to alleviate the burden on low-level lateral and longitudinal controllers, a finer discretization of the action space is employed, introducing intermediate lateral and speed references to improve trajectory tracking. The main contributions are summarized as follows.
\begin{itemize}
    \item We present DecisionPerceiver, an attention-based behavioral planning architecture that maps dynamic-size inputs into a compact latent space, reducing computational complexity while enabling scalable and finer control of the feature granularity.
    \item A finer discretization of the action space is introduced to reduce low-level control effort while preserving real-time planner inference performance.
\end{itemize}
The proposed approach is extensively benchmarked against state-of-the-art methods across multiple representative test scenarios. Quantitative evaluations are presented to demonstrate superior velocity tracking performance, reflected in improved speed regulation accuracy and smoother longitudinal profiles. The results consistently indicate enhanced dynamic feasibility and responsiveness under varying operational conditions.
\section{Methodology} \label{sec:method}
This section presents the proposed interaction-aware attention-based network for high-level behavioral planning. The formulation of the refined discrete action set is described, along with its integration into the decision-making framework. The reinforcement learning training environment and optimization procedure are also outlined.
\subsection{Actor-critic framework}
The high-level behavioral planning problem is formulated as a Markov Decision Process (MDP), in which an agent in state $s_t$ executes an action $a_t\sim\pi$ according to a policy $\pi$ at discrete time $t_k$. The action $a_t$ transitions the state to $s_{t+1}$ and the agent receives the reward signal $r_t$. The objective is to determine a policy that maximizes the expected return $R(s_t)=\sum_{t'\geq t}\gamma^{t'-t}r_{t'}$ where the discount factor $\gamma \in [0,1]$ reduces the influence of the later steps. Among the various solution methods, policy gradient algorithms are widely adopted. These methods estimate the gradient
\begin{equation}
 \hat{g}_t = \hat{\mathbb{E}}_t[\nabla_{\theta}\log\pi_{\theta}(a_t|s_t)\hat{A}_t],   
\end{equation}
for a parameterized policy $\pi_\theta$ and an estimate of the advantage function $\hat{A}_t$, which measures the relative benefit of executing action $a_t$ in state $s_t$ compared to the expected value under the current policy. This gradient estimator faces two challenges: \textit{i)} high variance in the advantage estimate $\hat{A}_t$, and \textit{ii)} instability from unconstrained updates. The variance is mitigated by estimating the advantage with a learned state-value function $\hat{V}_\theta(s_t)$. While trust-region constraints based on KL-divergence can be introduced, practical alternatives exist. Proximal Policy Optimization (PPO) \cite{schulman2017proximalpolicyoptimizationalgorithms} employs a clipped surrogate objective.
\begin{equation}
    L^{\text{CLIP}}(\theta) = \mathbb{E}[\min(\eta_t(\theta)\hat{A}_t,\text{clip}(\eta_t(\theta),1-\epsilon,1+\epsilon)\hat{A}_t)]
\end{equation}
where $\eta_t(\theta)=\frac{\pi_{\theta}(a_t|s_t)}{\pi_{\theta_{\text{old}}}(a_t|s_t)}$ denotes the policy ratio. The advantage term  $\hat{A}_t$ reduces gradient variance and weights updates according to relative performance (i.e., if the trajectory surpasses the expected reward), while the $\min$ operator enforces a pessimistic bound and the $\text{clip}$ operator enforces updates within the interval $[1-\epsilon,1+\epsilon]$. 

In combination with the value function squared-error loss $L^V = (\hat{V}_\theta(s_t)-V_t^\text{target})^2$ and an entropy regularization term $S[\pi_\theta](s_t)$ to promote exploration, the objective is defined as
\begin{equation}
    L^{\text{PPO}}_t(\theta) = \hat{\mathbb{E}}_t[L^{\text{CLIP}}(\theta) - c_1L^V(\theta) + c_2S[\pi_\theta](s_t)],
\end{equation}
with weighting coefficients $c_1$ and $c_2$. The PPO algorithm samples trajectories of length $T$ using the behavior policy $\pi_{\theta_{\text{old}}}$ and estimates the advantage at each step via Generalized Advantage Estimation \cite{schulman2018highdimensionalcontinuouscontrolusing}. The objective $L^{\text{PPO}}$ is then optimized with respect to $\theta$ for $K$ epochs using gradient-based optimizers such as SGD or Adam \cite{ruder2017overviewgradientdescentoptimization}.
\subsection{Interaction-aware attention-based network}
\begin{figure}[t!]
    \centering
    \includegraphics[width=0.86\linewidth]{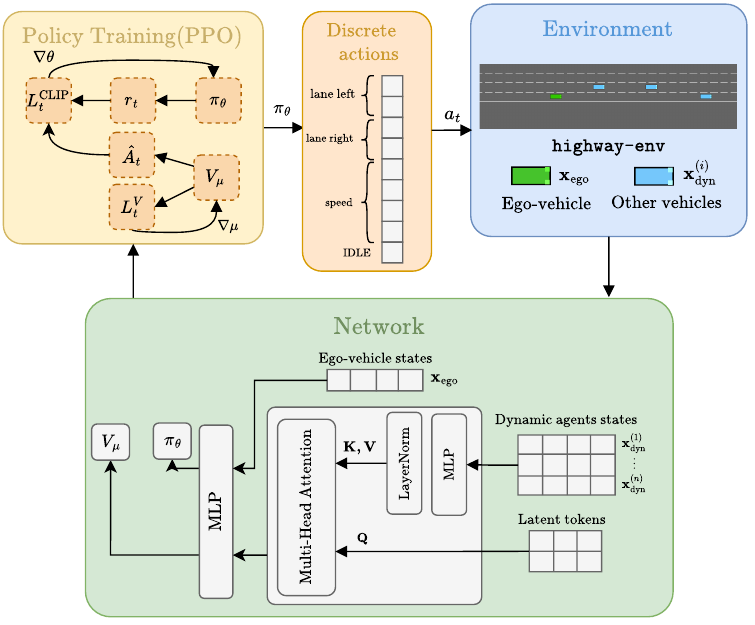}
    \caption{DecisionPerceiver architecture receives ego-vehicle and surrounding agents’ states as static/dynamic inputs to a multi-head attention module. The network is optimized within an actor-critic framework by back-propagating gradients through both the policy and value functions. The environment executes actions sampled from the policy $\pi_{\theta}$ based on current observations, yielding the finer discrete action set $\mathcal{A}_{\text{new}}$.} 
    \label{fig:method}
\end{figure}

DecisionPerceiver is proposed in this paper as an interaction-aware architecture that models dependencies between the ego vehicle and surrounding agents using multi-head attention, naturally accommodating variable-size inputs. The attention mechanism evaluates pairwise interactions between queries $\mathbf{Q}$ and keys $\mathbf{K}$, scales the similarity scores by $1/\sqrt{d}$, and applies the resulting weights to the values $\mathbf{V}$:
\begin{equation}
    \text{Attention}(\mathbf{Q,K,V}) = \text{softmax}\left(\frac{\mathbf{Q}\mathbf{K}^\top}{\sqrt{d}}\right)\mathbf{V}
\end{equation}
Further details are provided in \cite{vaswani2023attentionneed}. Conventional attention scales quadratically with input size, as each token attends to all others, and offers limited control over feature granularity. Inspired by Perceiver IO \cite{jaegle_perceiver_2022} and Wayformer \cite{nayakanti2022wayformermotionforecastingsimple}, the inputs are projected into a latent space $\mathbf{z}\in\mathbb{R}^{N\times{}D}$, where $N$ denotes the number of latent tokens. This latent bottleneck bounds computational complexity and enables controllable feature abstraction (see Fig.~\ref{fig:method}). While early feature projection reduces the ability to extract attention maps for interpretability, the experimental results indicate that this trade-off leads to overall performance improvement. 

In the proposed framework, the dynamic agent states $\{\mathbf{x}^{i}_\text{dyn}\}_{i=1:M}\in\mathbf{R}^{M\times{}F}$ are encoded using a multi-layer perceptron (MLP) followed by Layer Normalization. The resulting features serve as keys $\mathbf{K}$ and values $\mathbf{V}$, while attention scores are computed with respect to latent token queries $\mathbf{Q}$. This preprocessing stage is model-agnostic and may be replaced by alternative feature extraction schemes, such as learned Fourier feature mappings \cite{schier_deep_2023}.

The encoded dynamic features (i.e., interaction-aware latent representation) are fused with the fixed-size ego state $\mathbf{x}_{\text{ego}} \in \mathbb{R}^{1\times F}$ through an additional MLP layer, enabling joint reasoning over self and surrounding agents. From this unified embedding, separate network heads regress the value function $V_\theta$ and the policy $\pi_\theta$ under an actor–critic framework \cite{NIPS1999_6449f44a}. Both heads are optimized via backpropagation using the clipped PPO objective $L^{\text{CLIP}}$.
Throughout the network, no permutation-invariant collapsing operator, such as pooling, is employed, since such operations may hinder learning of inter-agent interactions, unlike in DeepSet models.
\subsection{Discrete action set}
As another contribution of this paper, and in order to reduce the control effort required by the longitudinal and lateral low-level controllers, the standard discrete action set for behavioral planning, $\mathcal{A} = $\{\texttt{keep lane} , \texttt{turn left}, \texttt{turn right}, \texttt{faster}, \texttt{slower}\} \cite{highway-env,hugle_dynamic_2019}, is extended with a finer discretization of steering and velocity-related actions (e.g $\texttt{turn left} \xrightarrow{}$ \{\texttt{turn half left,  turn full left}\}). This refinement reduces the control effort by the low-level controllers that track high-level meta-actions. All actions, except $a_t=\texttt{keep lane}$, are discretized into two levels, yielding an augmented action space $\mathcal{A}_{\text{new}}$ comprising nine discrete meta-actions. More discretization bins have improved performance in RL tasks \cite{tang2020discretizingcontinuousactionspace} by introducing inductive bias, reducing gradient variance, and creating a more compact search space. Furthermore, for lower-level controllers, regularization control with a discrete reference set is easier to monitor than tracking control with a continuous reference.
\subsection{Observations} 
In the developed framework, network observations are divided into static and dynamic inputs. The ego-vehicle state $\mathbf{x}_{\text{ego}} = [x,~y,~v_x,~v_y,\cos\theta,\sin\theta] \in \mathbb{R}^6$ constitutes the static input, where $(x,y)$ denote Cartesian position, $v_x,v_y$ are velocities in the world frame, and $(\cos(\theta),\sin(\theta))$ provide an orientation parameterization that avoids angle wrapping. The set of surrounding agent states $\{\mathbf{x}^{i}_\text{dyn}\}_{i=1:M}$ forms the dynamic input, with $\mathbf{x}^{i}_{\text{dyn}}=[\Delta{}x_i,\Delta{}y_i,\Delta{}v_{x,i},\Delta{}v_{y,i},\cos\Delta{}\theta_i, \sin\Delta{}\theta_i]$. Here, ($\Delta{}x,\Delta{}y$), $(\Delta{}v_x,\Delta{}v_y)$, and $(\cos\Delta{}\theta, \sin\Delta{}\theta)$ denote relative position, velocity, and orientation (of the agent) with respect to the ego vehicle. This relative encoding improves robustness to variations in scenario scale.
\subsection{Training environments}
The proposed method and baselines are evaluated across multiple RL scenarios in the \textit{highway-env} simulator \cite{highway-env}, with diverse road topologies and interaction dynamics (Fig.~\ref{fig:snapshot}). Our simulation choice is based on quick deployment and replicability of environments as well as low footprint for parallel training. 
\subsubsection{Highway}
The objective is to drive on a multi-lane highway at a target speed $v_{\text{goal}}$ while avoiding collisions. Driving in the right lane is encouraged due to higher speed limits. The reward consists of a scaled velocity term $r_v=k_v\frac{v}{v_{\text{max}}},~k_v=0.4$, a collision penalty, and a right-lane incentive.
\begin{equation*}
r( s,a) =\mathds{1}_{\text{on-road}}*( r_{v} -\mathds{1}_{\text{collision}} +\mathds{1}_{\text{right}}),    
\end{equation*}
\subsubsection{Intersection}
This scenario requires negotiating surrounding traffic to safely execute an unprotected left turn and reach the goal lane. While maintaining the target speed is desirable, goal attainment is prioritized. The reward comprises velocity, collision-avoidance, and goal-reaching terms.
\begin{equation*}
r( s,a) =\mathds{1}_{\text{on-road}}*( r_{v} -\mathds{1}_{\text{collision}} +\mathds{1}_{\text{goal}})
\end{equation*}
\subsubsection{Roundabout}
The agent must enter the roundabout, interact with traffic, and exit at the designated goal. A reference path is predefined, emphasizing lane changes, reaching $v_{\text{max}}$, and collision avoidance. The reward includes velocity, collision-penalty, and lane-change incentive terms.
\begin{equation*}
r( s,a) =\mathds{1}_{\text{on-road}}*( r_{v} -\mathds{1}_{\text{collision}} +\mathds{1}_{\text{lane}}).  
\end{equation*}
\section{Simulation Results and Discussions} \label{sec:benchmarking}
This section presents the training procedure for the proposed network and benchmarking results against prominent baselines. Comprehensive ablation studies are also conducted to assess the contribution of key architectural components and action-space refinements.
\subsection{Training procedure}
The proposed network employs an MLP with 128 neurons followed by a normalization layer to generate the keys $\mathbf{K}$ and values $\mathbf{V}$ for the multi-head attention. The baseline configuration uses latent tokens of size $(4,128)$, whose initial values are sampled from a uniform distribution $\mathcal{U}\sim{}[0,1]$. The multi-head attention layer consists of 4 heads and projection matrices of size $(128,128)$. In case of a dimensional mismatch between the latent tokens and the dynamic input, we apply the tiling operation over the latent tokens. The resulting attention scores are concatenated with the ego-state input $\mathbf{x}_{\text{ego}}$ and passed to two MLPs to model the policy and value functions, each with 128 neurons and ReLU nonlinearities.  

The novel architecture and baselines are trained using Stable Baselines3's PPO algorithm \cite{stable-baselines3} with 24 parallel agents to accelerate the rollout phase, for $1\times10^6$ time steps for the \textit{highway} and \textit{intersection} environments and $3\times10^6$ time steps for the \textit{roundabout} scenario, with 64 rollout steps and a batch size of 128 samples. The optimization process consists 10 epochs per update. We set the discount factor $\gamma=0.95$ to encourage action exploration, as initial experiments showed the optimization getting stuck in local minima, leading to a biased policy towards a single action. The same procedure was initially applied to the entropy coefficient $c_2$, but without further improvement; therefore, the default value of 0.0 was retained. We also tune the value network coefficient $c_1=0.25$, as we observe a correlation between improved value regression (i.e., lower value loss) and policies that yield higher returns. Every $1\times10^5$ steps, we evaluate the policies for 200 episodes to obtain intermediate results. After training, we evaluate over 500 episodes and report results as test metrics. We use DeepSet \cite{hugle_dynamic_2019}, LFFDS \cite{schier_learned_2023}, and Ego-attention \cite{leurent_social_2019} as baselines because they are publicly available.

\begin{figure*}[t!]
    \centering
    \begin{subfigure}[t]{0.329\textwidth}
        \centering
        \includegraphics[width=0.95\textwidth]{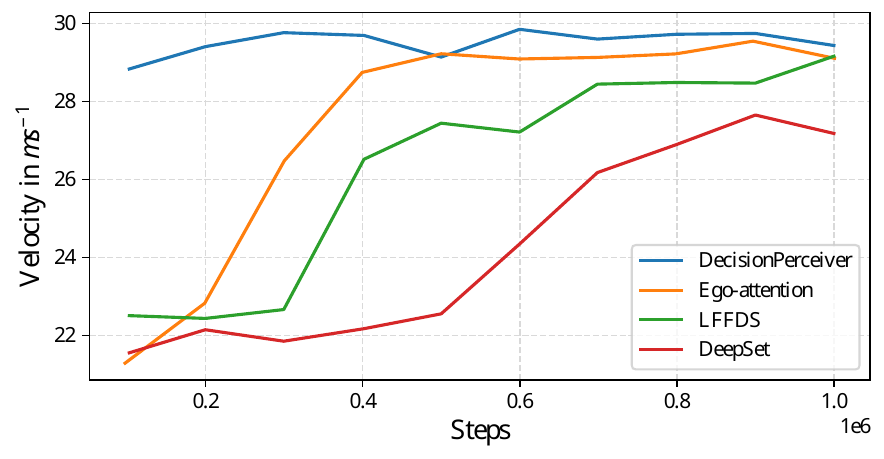}
        \caption{\textit{Highway-v0}}
    \end{subfigure}
    \hfill
    \begin{subfigure}[t]{0.329\textwidth}
        \centering
        \includegraphics[width=0.95\textwidth]{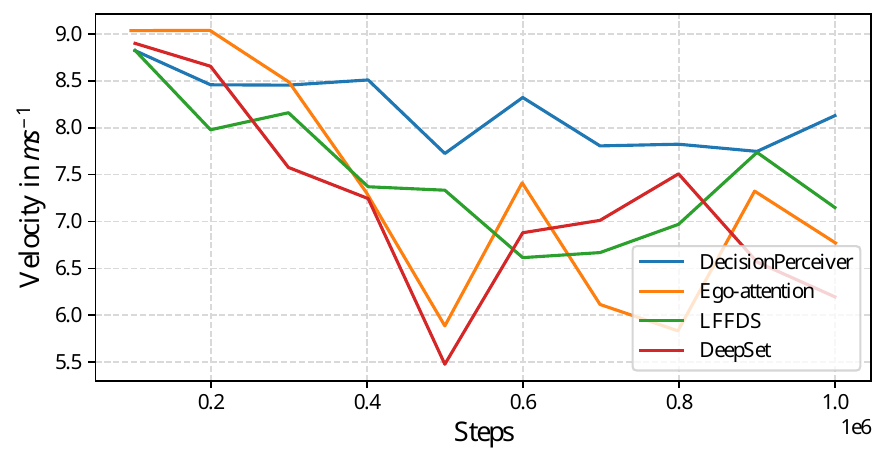}
        \caption{\textit{Intersection-v0}}
    \end{subfigure}
    \hfill
    \begin{subfigure}[t]{0.329\textwidth}
        \centering
        \includegraphics[width=0.95\textwidth]{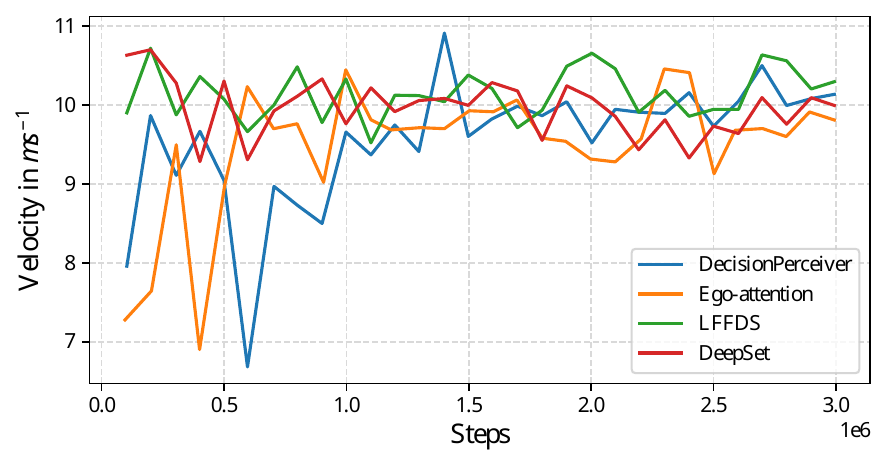}
        \caption{\textit{Roundabout-v0}}
    \end{subfigure}
    
    \caption{Average velocity over 200 evaluations runs during the training process on the three driving environments.}
    \label{fig:training_plot}
\end{figure*}
\subsection{Benchmarking}
As shown in Fig.~\ref{fig:training_plot}, DecisionPerceiver achieves competitive velocity performance relative to state-of-the-art methods, with evaluations conducted every $1\times10^5$ training steps
\subsubsection{Training evaluation}
In \textit{Highway}, DecisionPerceiver achieves a steady average velocity of 28.5 \si{\meter\per\second} from the first evaluation at $1\times10^{5}$ steps until convergence, indicating that interaction modeling accelerates learning in overtaking scenarios. Ego-attention requires $5\times10^{5}$ steps to reach comparable performance. This highlights the advantage of projecting agent interactions into a fixed-size latent space, although both architectures employ attention backbones. In contrast, DeepSet and its variant LFFDS exhibit slower convergence than Ego-attention, resulting in lower speed performance.

In \textit{Intersection}, interaction modeling is critical for avoiding collisions during unprotected left turns while limiting unnecessary speed reductions. DecisionPerceiver achieves the highest final average velocity (8.25\si{\meter\per\second}), followed by LFFDS (7.4\si{\meter\per\second}), Ego-attention, and DeepSet. All methods initially favor aggressive maneuvers but progressively learn to balance speed, safety, and traffic flow efficiency. DecisionPerceiver maintains the strongest high-speed performance among them.

\begin{figure*}[t!]
    \centering
    \begin{subfigure}[t]{0.32\textwidth}
        \centering
        \includegraphics[width=1.0\textwidth]{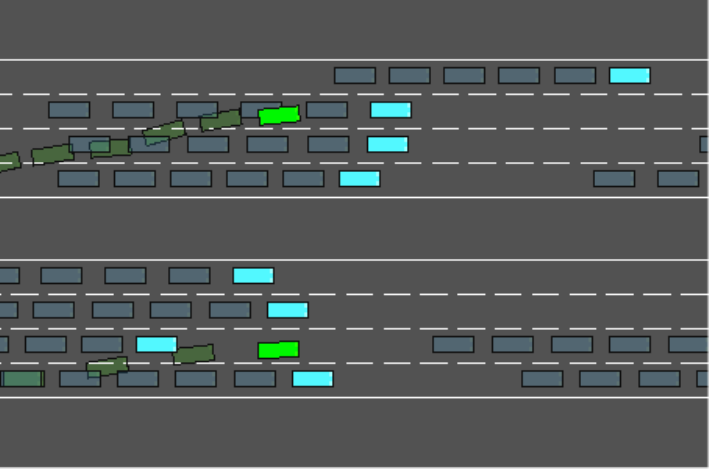}
        \caption{\textit{Highway-v0}}
    \end{subfigure}
    \hfill
    \begin{subfigure}[t]{0.32\textwidth}
        \centering
        \includegraphics[width=0.61\textwidth]{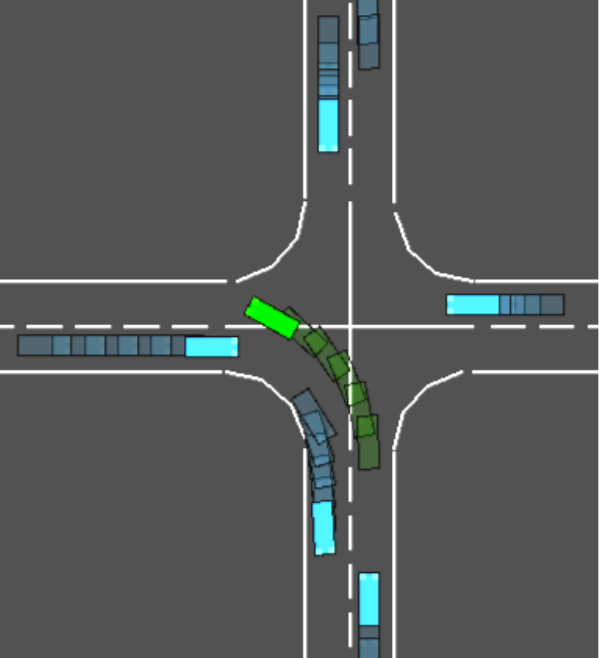}
        \caption{\textit{Intersection-v0}}
    \end{subfigure}
    \hfill
    \begin{subfigure}[t]{0.32\textwidth}
        \centering
        \includegraphics[width=0.71\textwidth]{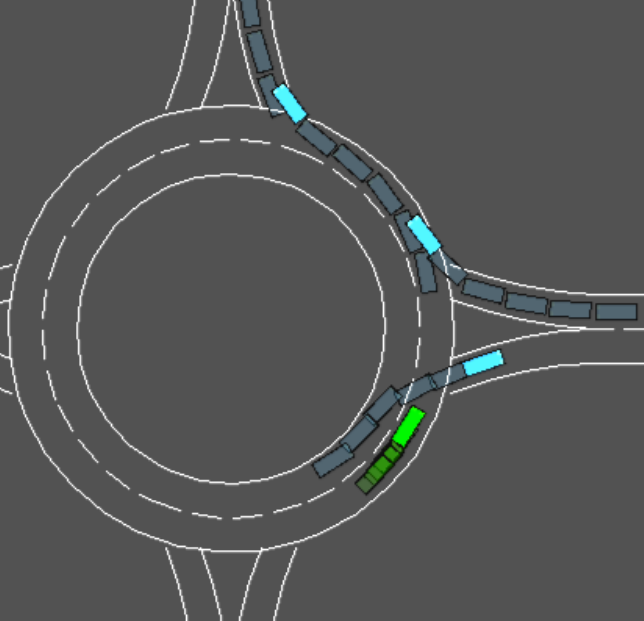}
        \caption{\textit{Roundabout-v0}}
    \end{subfigure}
    
    \caption{Motion trajectory snapshots in the traffic environments from \textit{highway-env} \cite{highway-env} simulator using actions sampled from \textit{DecisionPerceiver} after training. The interaction awareness enhances the reaction towards conflicting trajectories with the ego trajectory by taking counter measures such as accelerating to overtake traffic flow (such as in \textit{Highway} and \textit{Intersection}) or braking to wait for the following agent to merge onto its target lane (see trajectory of \textit{Roundabout}).}
    \label{fig:snapshot}
\end{figure*}

In \textit{Roundabout}, the interaction-aware methods Ego-attention and DecisionPerceiver exhibit similar training curves, reaching a steady state at $1.5\times10^{6}$ steps. DeepSet and LFFDS maintain nearly constant velocity throughout training. As this scenario involves limited interaction (typically only two nearby vehicles), attention-based models show slower convergence due to more expressive scene modeling, whereas set-based methods benefit from simpler representations. Despite slower learning, DecisionPerceiver and Ego-attention achieve the highest and second-highest velocities at $3.0\times10^{6}$ steps, highlighting the value of interaction awareness. Training visualizations for DecisionPerceiver are shown in Fig.~\ref{fig:snapshot}.
\subsubsection{Test performance}
The test metrics in Tables~\ref{tab:highway}-\ref{tab:roundabout} complement the training results. In \textit{Highway}, DecisionPerceiver and Ego-attention rank first and second across all three metrics. Relative to DeepSet and LFFDS, DecisionPerceiver achieves gains of 6.247\% in velocity, 4.203\% in reward, and 24.1\% in early termination rate. These results confirm robust performance across varied traffic conditions, with occasional suboptimal rankings arising in scenarios that do not fully exploit its interaction-awareness modeling capabilities. 

In \textit{Intersection}, DecisionPerceiver attains the highest average speed (8.243\si{\meter\per\second}), followed by LFFDS (7.137\si{\meter\per\second}), yielding a 15.4\% relative gain. Ego-attention achieves higher reward and the lowest early termination rate, whereas DecisionPerceiver maintains a more aggressive high-speed policy at the expense of safety.

In \textit{Roundabout}, DecisionPerceiver achieves a velocity of 9.951\si{\meter\per\second}, which is 5\% and 1\% lower than the results of DeepSet and LFFDS, respectively. However, it records a lower early termination rate than both methods (approximately 5$\times$ less) and a higher reward score, indicating that the resulting policy is comparatively safer. 
\begin{table}[!h]
\centering
\caption{Average metrics of velocity, reward and early termination rate over 500 test runs on \textbf{Highway} environment. Best results are highlighted in \textbf{bold} and second best are \underline{underlined}.}
\resizebox{0.45\textwidth}{!}{
        \begin{tabular}{lccc}
        \toprule
        \textbf{Model} & \textbf{Velocity}[\si{\meter\per\second}] \ $\uparrow$ & \textbf{Reward} \ $\uparrow$ & \textbf{Early.term}[\%] \ $\downarrow$  \\
        \midrule
        DeepSet & 27.53$\pm$2.233 & 32.69$\pm$6.012 & \underline{8.800}$\pm$2.830 \\
        LFFDS & 28.07$\pm$3.231 & 32.70$\pm$6.147 & 18.02$\pm$4.098 \\
        Ego-attention & \underline{29.08}$\pm$1.538 & \textbf{36.31}$\pm$6.404 & \textbf{0.000}$\pm$0.000 \\
        DecisionPerceiver & \textbf{29.25}$\pm$1.556 & \underline{34.66}$\pm$7.769 & \textbf{0.000}$\pm$0.000 \\
        \bottomrule
    \end{tabular}
}
\label{tab:highway}
\end{table}
 \begin{table}[!h]
\centering
\caption{Average metrics of velocity, reward and early termination rate over 500 test runs on \textbf{Intersection} environment. Best results are highlighted in \textbf{bold} and second best are \underline{underlined}.}
\resizebox{0.45\textwidth}{!}{
        \begin{tabular}{
            lccc}
        \toprule
        \textbf{Model} & \textbf{Velocity}[\si{\meter\per\second}] \ $\uparrow$ & \textbf{Reward} \ $\uparrow$ & \textbf{Early.term}[\%] \ $\downarrow$  \\
        \midrule
        DeepSet & 7.014$\pm$2.410 & 4.196$\pm$1.697 & \underline{32.60}$\pm$15.95 \\
        LFFDS & \underline{7.137}$\pm$2.384 & \underline{4.330}$\pm$1.571 & 33.20$\pm$17.80 \\
        Ego-attention & 6.881$\pm$2.537 & \textbf{4.547}$\pm$1.434 & \textbf{27.40}$\pm$14.60 \\
        DecisionPerceiver & \textbf{8.243}$\pm$1.788 & 4.238$\pm$1.826 & 43.20$\pm$19.50 \\
        \bottomrule
    \end{tabular}
}
\label{tab:intersection}
\end{table}
 \begin{table}[!h]
\centering
\caption{Average metrics of velocity, reward and early termination rate over 500 test runs on \textbf{Roundabout} environment. Best results are highlighted in \textbf{bold} and second best are \underline{underlined}.}
\resizebox{0.45\textwidth}{!}{
        \begin{tabular}{
            lccc}
        \toprule
        \textbf{Model} & \textbf{Velocity}[\si{\meter\per\second}] \ $\uparrow$ & \textbf{Reward} \ $\uparrow$ & \textbf{Early.term}[\%] \ $\downarrow$ \\
        \midrule
        DeepSet & 9.771$\pm$3.061 & 17.90$\pm$2.623 & 5.400$\pm$2.260 \\
        LFFDS & \textbf{10.07}$\pm$2.814 & 17.93$\pm$2.081 & \underline{5.200}$\pm$2.202 \\
        Ego-attention & 9.326$\pm$2.592 & \textbf{18.28}$\pm$1.332 & \textbf{0.000}$\pm$0.000 \\
        DecisionPerceiver & \underline{9.951}$\pm$2.922 & \underline{18.05}$\pm$1.960 & \textbf{0.000}$\pm$0.000 \\
        \bottomrule
    \end{tabular}
}
\label{tab:roundabout}
\end{table}
\subsection{Ablation studies}
\subsubsection{Introduction of new discrete set}
\begin{figure}[htpb!]
    \centering
    \includegraphics[width=0.825\linewidth]{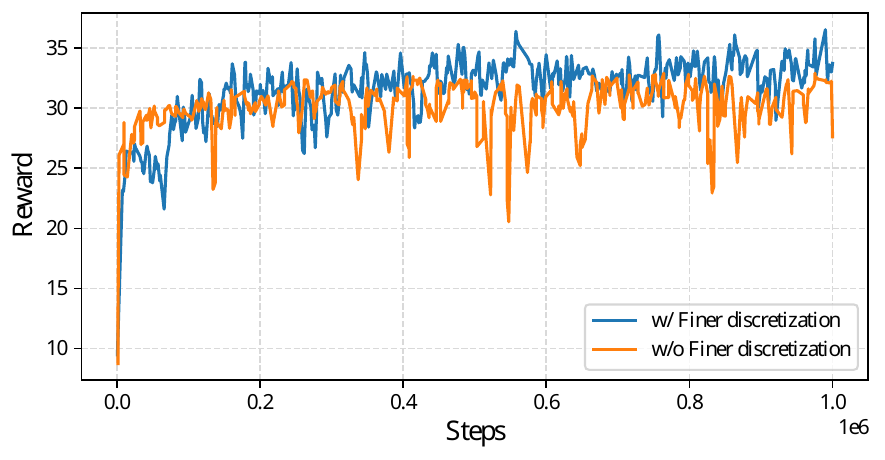}
    \caption{Training reward results in \textit{Highway} scenario with and without finer discretization.}
    \label{fig:reward_discrete}
\end{figure}

To better understand the impact of the design choices of the presented method, we analyze the performance gain from introducing a finer discretization of the action set, as well as the performance degradation that occurs as the number of vehicles within the perceived range increases. We select the \textit{Highway} scenario as the testbed since it provides a necessary condition under which $\mathcal{A}_{\text{new}}$ can be fully exploited: several adjacent driving lanes where the intermediate lane transitions can be accomplished more smoothly with the new discrete action set. In contrast, in \textit{Intersection} and \textit{Roundabout}, encouraging only finer speed control has limited effect because lane change improvement is constrained by fewer lanes. Figure~\ref{fig:reward_discrete} illustrates how the finer discretization increases the accumulated reward relative to the original action set after $1\times10^5$ steps and maintains a steady improvement until the end of the training. Meanwhile, the original action set exhibits several downward spikes every $2\times10^5$ steps and achieves a lower final value, indicating a more unstable training process. This behavior results directly from smaller reference changes in the low-level controller reducing reaction time and, consequently, the likelihood of collisions. Over time, this translate into a higher average speed and a higher accumulated reward. 

\subsubsection{Input scalability}
Perception scalability is evaluated by progressively increasing traffic density in the \textit{Highway} scenario and the maximum number of surrounding vehicles included in the dynamic input $\mathbf{x}_{\text{dyn}}$, with the policy retrained from scratch for each setting. Performance is assessed via average speed over 200 test episodes for detection limits of 5, 10, 15, and 20 vehicles within sensing range. Although these limits specify the maximum observable vehicles, the actual number encoded in $\mathbf{x}_{\text{dyn}}$ varies due to traffic distribution. Increasing the agent limit by five results in a 15\%–30\% reduction in average speed, mainly due to reduced maneuverability and higher collision risk. Nevertheless, stable and competitive performance is maintained across all configurations, demonstrating scalability to denser traffic without catastrophic degradation

\section{Conclusions}\label{sec:conclusions}
In this work, DecisionPerceiver, a novel deep learning architecture for high-level behavioral planning in autonomous driving, is introduced. The attention-based design projects inter-agent interactions into a compact latent space, enabling complex interaction modeling while bounding computational and memory costs to the latent dimensionality. Unlike vanilla attention, the architecture accommodates dynamic-sized inputs without quadratic complexity growth. A finer discretization of the high-level action space is also introduced to refine lane-change and speed-control commands, reducing the burden on low-level controllers while preserving fast planner inference.

The effectiveness of the proposed approach is validated through extensive experiments across diverse interactive driving scenarios, including intersection negotiation, overtaking, and high-speed traffic navigation. Under identical training settings, the method consistently outperforms state-of-the-art attention-based and set-based baselines, including Ego-attention, DeepSet, and LFFDS, in terms of achieved average velocity. The proposed method achieves up to 15\% relative gain in average velocity and a 5-8 times lower collision rate than the strongest baseline in two of the three scenarios. These results demonstrate improved scalability and representation flexibility, yielding superior performance in complex traffic environments.
\appendices


\bibliographystyle{IEEEtran}
\bibliography{Refs}

\end{document}